# Building a comprehensive syntactic and semantic corpus of Chinese clinical texts

Bin He [a], Bin Dong [b], Yi Guan [a,*], Jinfeng Yang [c], Zhipeng Jiang [a], Qiubin Yu [d], Jianyi Cheng [a], Chunyan Qu [a]

[a] School of Computer Science and Technology, Harbin Institute of Technology, Harbin, China

[b] Ricoh Software Research Center (Beijing), Beijing, China

[c] School of Software, Harbin University of Science and Technology, Harbin, China

[d] Medical Records Room, the Second Affiliated Hospital of Harbin Medical University, Harbin, China

* Corresponding author

**Abstract**

*Objective:* To build a comprehensive corpus covering syntactic and semantic annotations of Chinese clinical texts with corresponding annotation guidelines and methods as well as to develop tools trained on the annotated corpus, which supplies baselines for research on Chinese texts in the clinical domain.

*Materials and methods:* An iterative annotation method was proposed to train annotators and to develop annotation guidelines. Then, by using annotation quality assurance measures, a comprehensive corpus was built, containing annotations of part-of-speech (POS) tags, syntactic tags, entities, assertions, and relations. Inter-annotator agreement (IAA) was calculated to evaluate the annotation quality and a Chinese clinical text processing and information extraction system (CCTPIES) was developed based on our annotated corpus.

*Results:* The syntactic corpus consists of 138 Chinese clinical documents with 47,424 tokens and 2553 full parsing trees, while the semantic corpus includes 992 documents that annotated 39,511 entities with their assertions and 7695 relations. IAA evaluation shows that this comprehensive corpus is of good quality, and the system modules are effective.

*Discussion*: The annotated corpus makes a considerable contribution to natural language processing (NLP) research into Chinese texts in the clinical domain. However, this corpus has a number of limitations. Some additional types of clinical text should be introduced to improve corpus coverage and active learning methods should be utilized to promote annotation efficiency.

*Conclusions:* In this study, several annotation guidelines and an annotation method for Chinese clinical texts were proposed, and a comprehensive corpus with its NLP modules were constructed, providing a foundation for further study of applying NLP techniques to Chinese texts in the clinical domain.

**1. Introduction**

Electronic medical records (EMRs) represent the storage of all healthcare data and information in electronic formats [1] and constitute core data in the implementation of health care services. These services are undergoing enormous changes with increasing health awareness and demand for medical services. The situation is becoming more urgent for China, a country with the largest population but limited medical resources. In facing these challenges, the Chinese

Ministry of Health (MOH) has issued a series of relevant regulations since 2010 to standardize EMR systems and their intelligent support [2-4]. With the rapid popularization of EMRs, the development of healthcare services has a solid data foundation.

Clinical texts, an important type of patient data within EMRs, are free-text documents that contain large amounts of information about patients' medical activities. In recent years, natural language processing (NLP) techniques on English clinical texts have been widely used [5, 6] and many resources have been established for the development of these techniques. For example, the Unified Medical Language System (UMLS) [7], an integrated knowledge base of biomedical concepts, is widely applied in medical informatics research. Moreover, challenges organized by Informatics for Integrating Biology & the Bedside (i2b2) have released various kinds of annotated data for medical information extraction tasks, and enable clinical researchers to employ these clinical corpora for discovery research [8].

However, due to the lack of an annotated corpus, NLP research on Chinese clinical texts is still at a preliminary stage. Chinese clinical text has sublanguage features [9] that make it difficult for research on general-domain texts to be applied directly to clinical texts. In this study, we focus our efforts on conducting syntactic and semantic annotations of Chinese clinical texts, involving two resident physicians (P1 and P2) and eight annotators with backgrounds in computational linguistics (CL1-CL10). To our knowledge, this is the first comprehensive Chinese clinical corpus that includes several types of syntactic and semantic annotations, making it possible to develop effective NLP techniques for application to Chinese texts in the clinical domain.

This paper has six sections and is organized as follows: background on NLP research on clinical texts is summarized in Section 2. We then describe the development of annotation guidelines, annotation method, and annotation quality measurement in Section 3. Next, Section 4 presents inter-annotator agreement (IAA) scores, data analysis of the annotations, and system development based on this corpus. In Section 5, we describe the contributions of this work and identify further improvements for future work.

**2. Background**

NLP tasks can be divided into low-level tasks and higher-level tasks: low-level tasks include sentence boundary detection, tokenization, word segmentation, part-of-speech (POS) tagging, shallow parsing, and so on; based on low-level tasks, higher-level tasks include named entity recognition (NER), negation identification, relationship extraction, etc. [10] As a bridge for adapting existing techniques into the clinical domain, annotated corpora in the clinical domain are needed. Table 1 summarizes some major clinical text corpora for NLP tasks, and is discussed in the following sub-sections.

**Table 1**

Clinical text corpora for research on low-level and higher-level NLP tasks

| Part A | | | | | | | |
|---|---|---|---|---|---|---|---|
| Author | Year | Language | Scale | Chinese word segmentation | POS tagging | Shallow parsing | Full parsing |
| Savova *et al.* [6] | 2010 | English | 273 documents | – | √ | √ | – |
| Albright *et al.* [11] | 2013 | English | 13,091 sentences | – | √ | √ | √ |
| Fan *et al.* [12] | 2013 | English | 1100 sentences | – | √ | √ | √ |
| Xu *et al.* [13] | 2014 | Chinese | 336 documents | √ | – | – | – |

| Zhang et al. [14] | 2016 | Chinese | 100 documents | √ | – | – | – |
| Part B | | | | | | | |
| Author | Year | Language | Scale | Entities | Assertions | Relations | |
| Meystre et al. [15] | 2006 | English | 160 documents | √ | √ | – | |
| Roberts et al. [16] | 2009 | English | 150 documents | √ | √ | √ | |
| Savova et al. [6] | 2010 | English | 160 documents | √ | √ | – | |
| Uzuner et al. [17] | 2011 | English | 826 documents | √ | √ | √ | |
| Albright et al. [11] | 2013 | English | 13,091 sentences | √ | √ | – | |
| Elhadad et al. [18] | 2015 | English | 531 documents | √ | √ | – | |
| Xu et al. [13] | 2014 | Chinese | 336 documents | √ | – | – | |
| Lei et al. [19] | 2014 | Chinese | 800 documents | √ | – | – | |
| Wang et al. [20] | 2014 | Chinese | 11 613 CCs | √ | – | – | |
| Wang et al. [21] | 2014 | Chinese | 115 documents | √ | – | – | |
| Jia et al. [22] | 2014 | Chinese | 30 documents | √ | √ | – | |
| Xu et al. [23] | 2015 | Chinese | 500 HPIs | √ | √ | – | |
| Li et al. [24] | 2015 | Chinese | 1000 documents | √ | – | √ | |

"√" means annotated, and "–" means unannotated. POS, part-of-speech; CC, chief complaint; HPI, history of present illness.

*2.1. Annotated clinical text corpus for low-level tasks*

*2.1.1. Current status in English clinical texts*

The Mayo Clinic's cTAKES system aims at comprehensive processing of clinical texts and covers various NLP techniques [6]. In this work, a linguistic corpus annotated for POS tagging and shallow parsing was accomplished by three linguistic experts via extending the Penn TreeBank (PTB) annotation guidelines [25, 26] to the clinical domain. Additionally, Albright et al. [11] constructed a corpus involving annotations of POS tags and syntactic trees, and its advantage is that multilayer annotations are carried out in each sentence, which is beneficial in training joint models. As Albright et al. pointed out, the sentences in clinical texts contain numerous patterns that do not appear in the bracketing guidelines for the PTB [26], and clinical texts have sublanguage properties [27, 28]. Therefore, Fan et al. [12] developed annotation guidelines for parsing clinical texts and annotated a syntactic corpus of progress notes from the University of Pittsburgh Medical Center (UPMC).

*2.1.2. Current status in Chinese clinical texts*

Word segmentation is an initial processing step in low-level tasks on Chinese texts. Xu et al. [13] found that out-of-vocabulary words and resolving ambiguities in clinical texts brought great challenges to word segmentation and that a state-of-the-art Chinese word segmenter trained by a general corpus would have poor performance in the clinical domain. Therefore, they manually annotated a corpus of segmented words in discharge summaries to improve the performance of word segmenters in Chinese clinical texts. Analogously, Zhang et al. [14] constructed similar corpus to achieve better word embedding features.

*2.2. Annotated clinical text corpus for higher-level tasks*

*2.2.1. Current status in English clinical texts*

In 2006, Meystre *et al.* [15] constructed an entity corpus involving 80 different medical problems with their assertions to judge whether a medical problem is *present* or *absent*, and 10 clinical document types were annotated. However, this corpus was somewhat limited in that only medical problems and two kinds of entity assertions were annotated. To extract further information from clinical texts automatically, Roberts *et al.* [16] randomly chose 50 clinical narratives, 50 histopathology reports, and 50 imaging reports to annotate entities, relations, modifiers, co-references, and temporal information in the CLinical E-Science Framework (CLEF) project [29]. This was the first corpus that extended the number of entity types to six, and was the first attempt at annotating relations and temporal information in clinical texts. Moreover, an iterative approach was used to develop annotation guidelines, and this greatly inspired subsequent work to build high-quality corpora in the clinical domain. Besides, Savova *et al.* built a named entity corpus [6] that included disorder entities with attached UMLS concept unique identifiers (CUI) and assertions that are of the types *negated*, *current*, *history of*, *family history of*, and *possible*. This corpus has contributed towards the development of cTAKES system, which brings enormous benefits to subsequent clinical text studies. In 2010, Uzuner *et al.* [30] released a corpus that annotated concepts, assertions, and relations. Based on semantic types defined in UMLS, concepts are classified into medical problems, tests, and treatments; meanwhile, there are six types of assertions for medical problems and three groups of relations between concepts. Furthermore, the annotation guidelines [31-33] in this corpus are of great importance for corpus construction in the clinical domain. However, diseases and symptoms, which are treated differently in medical practice, are not subdivided in this corpus but are merged into medical problems. In fact, Uzuner *et al.* [34] split medical problems into diseases and symptoms in a study before the i2b2 2010 challenge. Considering the differences between disorders and symptoms in many medical applications, Albright *et al.* [11] annotated disorders as a semantic type independent of signs or symptoms, and built a corpus that annotated entities and their assertions. In 2015, to enhance NLP research in the clinical domain, Elhadad *et al.* [18] released a corpus that annotated disorders with various attributes in SemEval-2015 Task 14. The attributes of the disorders are beneficial for extracting deeper patient information in the clinical texts.

*2.2.2. Current status in Chinese clinical texts*

Referring to the concept annotation guidelines in the 2010 i2b2/VA challenge, Xu *et al.* [13] labeled medical problems, treatments, and tests in Chinese discharge summaries and added two more entity types, namely medication and anatomy. Medication is divided from treatment for further analysis on the usage and effectiveness of medications, and anatomy can help to locate positions of symptoms or tests. Similar to Xu *et al.*'s corpus on entities, Lei *et al.* [19] developed an entity corpus of discharge summaries and admission notes from Peking Union Medical College Hospital. Their entity categories differ from the 2010 i2b2/VA concept guidelines in that treatments are divided into procedures and medications. Moreover, Xu *et al.* [23] annotated medical terms on "history of present illness" section in clinical texts, and proposed an effective rule-based method. Differ from the above research on Chinese clinical texts, Jia *et al.* [22] manually marked up negated information of medical terms. To our knowledge, this is the first entity corpus with assertion information annotated in Chinese clinical texts.

Moreover, research into clinical texts of traditional Chinese medicine has gradually been taken into account. Wang *et al.* [20] conducted research in recognizing symptoms from the chief complaints, but the text types and entity categories in their corpus were relatively few. Li *et al.* [24] proposed a network-based correlation analysis method to detect the herb-symptom associations, and built a dataset of herb-symptom records that annotated correlations between symptoms and herbs. This study is meaningful to research of relation extraction from Chinese clinical texts.

However, for the clinical texts on a particular disease, the existing classification standards of medical entities are too rough, and some important information has not been distinguished effectively. In order to identify tumor-related information from Chinese operation notes, Wang *et al.* [21] manually annotated 12 entity types on operations, which revealed operation details and correlated strongly with patients' pathological status. This study provides a good reference for research on information extraction for specific diseases.

*2.3. Shortcomings of research on Chinese clinical texts*

Corpus construction on English clinical texts is a mature field, and its annotation scheme and evaluation method are of great significance for Chinese clinical texts. Considering the research status described above, research on Chinese clinical texts has three shortcomings: first, research on low-level tasks is quite limited, and this may cause performance improvement of higher-level tasks to encounter a bottleneck; second, as far as we know, only negated assertion and symptom-herb correlation have been annotated, other types of assertions or relations have not been annotated systematically; and third, guideline tuning and annotator training are needed in corpus construction, but descriptions of previous research efforts have not described these processing procedures. Based upon the above three points, along with the fact that no clinical corpus written in Chinese has been released to the public, it is imperative to build a comprehensive corpus that follows a complete annotation scheme.

By referring to the existing research on English clinical texts, we constructed a comprehensive corpus of Chinese clinical texts. In our annotation method, some existing well-developed guidelines were used and adapted into Chinese clinical texts in the process of annotation guideline tuning. Next, annotator training and various measures were conducted to ensure the quality of this corpus. Furthermore, according to the annotations in this corpus, corresponding automatic system modules were developed.

**3. Materials and methods**

*3.1. Types of clinical text*

Discharge summaries and progress notes employed in this work were randomly selected from clinical texts of the Second Affiliated Hospital of Harbin Medical University (a general hospital in China), and all identifying information was removed manually to protect patient privacy. These two types of clinical text are semi-structured documents, and free text in the document is divided into several sections, as listed in Fig. 1.

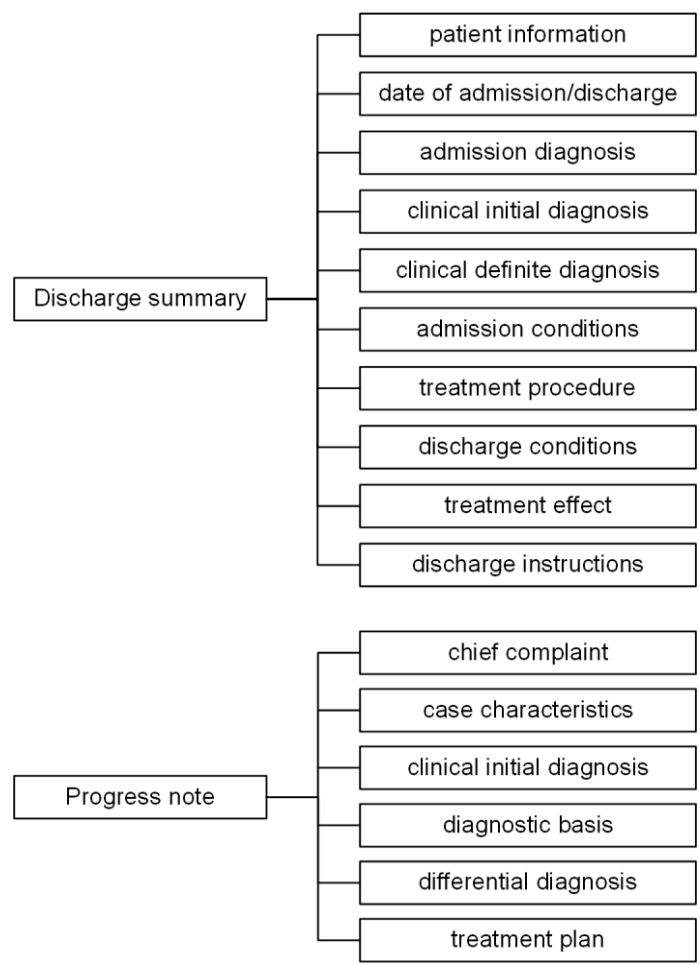

**Fig. 1.** Semi-structured sections in Chinese discharge summaries and progress notes.

*3.2. Annotation guidelines*

Due to the diversity of clinical texts, there is no existing annotation schema widely applicable in the clinical domain [11]. Owing to different language features between Chinese and English, annotation guidelines for CTB [35-37] were chosen to develop guidelines for low-level tasks on Chinese clinical texts; meanwhile, annotation guidelines in the 2010 i2b2/VA challenge [31-33] were consulted to develop guidelines for higher-level tasks. According to the characteristics of Chinese clinical texts, we developed several modified annotation guidelines [38-41] for four low-level and three higher-level tasks. Fig. 2 shows an example of the annotations in a sentence from the case characteristics section of a progress note.

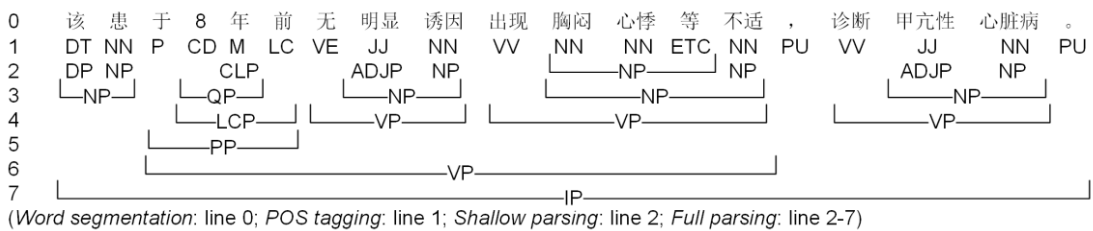

**Fig. 2.** An example of the annotations in a sentence from a progress note. NLP, natural language processing; DT, determiner; NN, common nouns; P, prepositions; CD, cardinal numbers; M, measure word; LC, localizer; VE, you3 as the main verb; JJ, noun-modifier other than nouns; VV, other verbs; ETC, tags for deng3 and deng3deng3 in coordination phrases; PU, punctuation; DP, determiner phrase; NP, noun phrase; CLP, classifier phrase; ADJP, adjective phrase; QP, quantifier phrase; LCP, phrase formed by "phrase + LC"; VP, verb phrase; PP, preposition phrase; IP, simple clause; POS, part-of-speech; SID, symptom indicates disease.

*3.2.1. Guideline development for low-level tasks*

*3.2.1.1. Word segmentation*

The segmentation guidelines for the Penn Chinese TreeBank (CTB) [35] cannot cover all the segmentation ambiguities in clinical texts, especially the segmentation of medical terms, abbreviations, and their combinations. In order to address these segmentation ambiguities, three word attributes were utilized in the word segmentation guidelines:

1. *Combinability* [42], which means that a word can be separated into two sub-words and that each sub-word has its independent POS;
2. *Reducibility*, which indicates that, if a word is an abbreviation, then it can be reverted to its complete expanded form to clarify its description;
3. *Replaceability*, which denotes that one sub-word in a combined word can be replaced by another word with the same POS, and that the new combined word may still appear in clinical texts.

Obviously, medical terms, if they are nouns or do not have the combinability attribute, are not normally split, such as "糖尿病 (diabetes)". Therefore, the main problems are segmentation ambiguities existing in non-nominal terms that have the combinability attribute. For these terms, a word segmentation method was developed, as shown in Fig. 3a, and Fig. 3b illustrates this method using the segmentation of "抗凝 (anti-coagulation)".

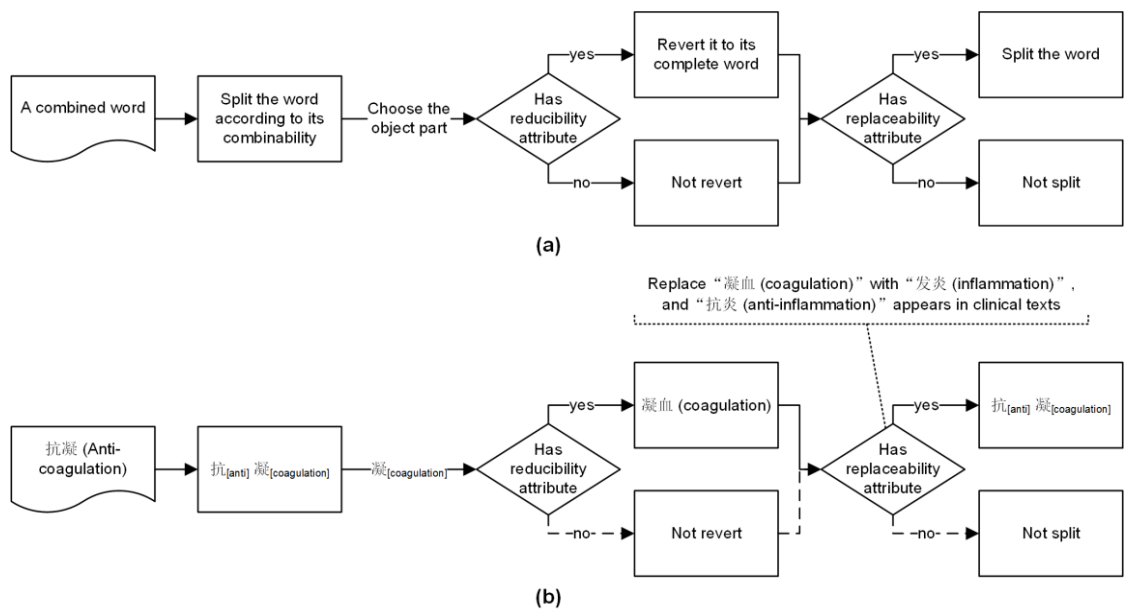

**Fig. 3.** A word segmentation method for non-nominal terms that have the combinability attribute.

*3.2.1.2. POS tagging and Parsing*

For the POS tagging task in Chinese clinical texts, the POS guidelines for CTB [36] also have some degree of incompleteness. Three main problems and their solutions are described as follows:

1. Some specific symbols that do not exist in CTB are used as abbreviations of certain words in clinical texts. For example, "+" in "肌力 4+级 (myodynamia level is 4+)" means "stronger". Moreover, we tagged the POS of the specific symbol based on its meaning in the context; hence, "+" in this example should be tagged as a VA (predicative adjective).
2. A verb-complement phrase is commonly utilized as an object to describe a patient's symptom in clinical texts, but this usage does not appear in CTB, so POS tags of words in a verb-complement phrase come with some ambiguity. For instance, "视物 模糊 (blurred vision)" in "伴有 视物 模糊 (with blurred vision)" is a symptom and can be seen as a noun phrase, so "视物[see things]" can be tagged as an NN (common nouns); but from the perspective of phrase structure, "视物[see things] 模糊[blurred]" is a verb-complement phrase, and thus "视物[see things]" should be tagged as a VV (other verbs). To solve this kind of ambiguity, POS tags are achieved according to the POS of the word itself, so "视物[see things]" is tagged as a VV.
3. Annotation ambiguities caused by omitting some parts of the sentence. For example, the POS tag of "左侧[left side]" in "左侧肢体麻木 (numbness in the left limbs)" and "右肺呼吸音清左侧弱 (right lung breath sounds clear and the left's weak)" are different. In CTB, ambiguity between "NN" and "JJ" can usually be disambiguated by judging whether the word is the head of a noun phrase; however, neither occurrence of "左侧[left side]" in the above two examples is the head of a noun phrase, so we need to complement omitted elements in the sentence. The former example has a normal grammatical structure in which "左侧[left side]" modifies "肢体

[limbs]" and should be tagged as a JJ; but "左侧[left side]" in the latter example is short for "左侧肺呼吸音 (left lung breath sounds)", which is a noun phrase and should be tagged as an NN.

Furthermore, we simplified the bracketing guidelines for CTB [37] and adapted these annotation specifications to the clinical domain, providing clear guidance in annotating for the parsing (shallow parsing and full parsing) task in Chinese clinical texts.

*3.2.2. Guideline development for higher-level tasks*

*3.2.2.1. Entities and assertions*

Concept annotation guidelines in the 2010 i2b2/VA challenge [31] include three categories of concept: medical problems, tests, and treatments. However, as Uzuner *et al.* [34] pointed out, patients' medical problems can be represented as diseases and symptoms, and these two kinds of concept have separate UMLS semantic types; hence, we treated diseases and symptoms as two types of medical entity in our annotation guidelines, as shown in Table 2A.

In the 2010 i2b2/VA challenge, only assertions of medical problems were annotated, and each medical problem was assigned one of the six assertion types [32]. In our work, we did not find any *hypothetical* entity in Chinese clinical texts, but observed a relatively frequent assertion in the default category *present*, so we deleted the *hypothetical* assertion type and separated the additional kind of assertion *occasional* from *present*. We assigned six assertion types to diseases and symptoms in Chinese clinical texts. Furthermore, because the statuses of treatments administered in patients are important references for clinical diagnoses, we annotated three types of assertion in treatments: *present*, *absent*, and *historical*. Table 2B lists the assertions of medical entities with their examples.

**Table 2**

Entities and their assertions annotated in Chinese clinical texts

| Part A | |
|---|---|
| Entity type | Example |
| Diseases | 行支气管镜检查示：小细胞肺癌 (Bronchoscopy showed: small cell lung cancer) |
| Symptoms | 疼痛时伴右下肢活动受限 (Pain accompanied by the right lower extremity activity limitation) |
| Tests | 行支气管镜检查示：小细胞肺癌 (Bronchoscopy showed: small cell lung cancer) |
| Treatments | 注射胰岛素控制血糖 (Injection of insulin to control blood glucose) |

| Part B | | | |
|---|---|---|---|
| Entity type | Assertion type | Description | Example |
| Diseases Symptoms | Present | Disease or symptom exists in the patient | 头 CT 示：双侧多发腔梗 (head CT **showed**: bilateral multiple lacunar infarct) |
| | Absent | Disease or symptom does not exist in the patient | 双下肢**无**浮肿 (**no** edema in both lower limbs) |
| | Possible | Disease or symptom may exist in the patient | 右肺下叶**考虑**创伤性湿肺 (Right lung lower lobe **consider** traumatic wet lung) |
| | Conditional | Disease or symptom occurs in the patient under certain conditions | ...胸闷、气短，常于**饮酒后**出现 (...chest tightness, shortness of breath, commonly occurs **after drinking**) |
| | Not associated with the patient | Disease or symptom exists in the patient's relatives | 患者**父母**均患有糖尿病 (**parents** of the patient suffer from diabetes) |
| | Occasional | Disease or symptom exists in the patient | **时有**胸闷气短 (there are chest tightness and |

| | | | occasionally | shortness of breath *sometimes*) |
|---|---|---|---|---|
| Treatments | Present | | The patient is experiencing or will experience the treatment | 右侧胸部见引流管 (drainage tube in the right side of the chest) |
| | Absent | | The patient does not experience the treatment | ***停用***达那唑 (***stop taking*** danazol) |
| | Historical | | The patient experienced the treatment in the past | *18 年前*剖宫产手术 (cesarean section *18 years ago*) |

In the examples, entities are underlined and indicators of the assertions are highlighted in bold and italics.

*3.2.2.2. Relations*

In Chinese clinical texts, entities of the same type usually appear one after the other in a sentence, and there are commonly concurrent relationships between these entities; for example, some treatments are administered for a disease or a disease causes some symptoms. Additionally, these entities may have the same type of relationship with an entity of a different type in the sentence, but one-to-one relationships between one of the former entities and the latter entity may not be clearly pointed out, causing some trouble in the annotation of one-to-one relationships. To avoid annotating fuzzy one-to-one relationships, we referred to the definition of narrative container that used in temporal relations [43], and proposed the concept of an "entity group" to assist in the relation annotation task in Chinese clinical texts.

Entities of the same type in a sentence are combined into an entity group if they satisfy the following two conditions: (1) simultaneity, which means that these entities appear at the same time during a medical activity of the patient, indicating a concurrent relationship between entities; (2) these entities have the same type of relationship with an entity of a different type in the sentence.

According to the definition of an entity group, one-to-one relationships can be developed into a relationship between an entity and an entity group, or a relationship between an entity group and another entity group. In the example shown in Fig. 2, the patient had symptoms of "胸闷 (chest congestion)" and "心悸 (palpitation)", and was diagnosed with "甲亢性心脏病 (hyperthyroid heart disease)", so a relationship between entity group "[胸闷；心悸]" and entity "甲亢性心脏病" was annotated.

The introduction of entity groups may weaken one-to-one relationships between entities, but solves the problem of fuzzy relationships. Besides, the definition of an entity group can also be explained by doctors' habits of clinical diagnosis and treatment: when a doctor makes a diagnosis based on the patient's current symptoms, the diagnosis is not based on one symptom but on a comprehensive judgment of a group of symptoms, and several tests or treatments are applied cooperatively to the patient.

In addition to the introduction of entity groups, we also made some adjustments to the relation types. Based on relations in the 2010 i2b2/VA challenge [33], we extended the relation types into five main categories and 16 subcategories in Chinese clinical texts, as shown in Table 3. All these relationships are bounded by sentences, and entity assertions are not considered when labeling relationships.

**Table 3**

Relations between medical entities annotated in Chinese clinical texts

| Entity pair | Relation | Description | Example |
|---|---|---|---|

|  | type |  |  |
|---|---|---|---|
| Treatments, Diseases | TrID | Treatment improves disease | ...诊断[贫血]_D，给予[输血]_Tr 后好转 (...was diagnosed with [anemia]_D, and improved after giving [blood transfusion]_Tr) |
|  | TrWD | Treatment worsen disease | [高血压病]_D 口服[替米沙坦]_Tr 控制，但血压控制不佳 (oral [Telmisartan]_Tr to control [hypertensive disease]_D, but poorly controlled blood pressure) |
|  | TrCD | Treatment causes disease | [电除颤]_Tr 后：[III度房室传导阻滞]_D (after [electric defibrillation]_Tr: [three degree atrioventricular block]_D) |
|  | TrAD | Treatment is administered for disease | ...被诊断为[结肠癌]_D，行[右半结肠癌根治术]_Tr (was diagnosed with [colon cancer]_D, and [right hemi-colonic carcinoma radical operation]_Tr was administered) |
| Treatments, Symptoms | TrIS | Treatment improves symptom | ...服用[钙剂]_Tr 等治疗后，[后背部疼痛]_S 显著缓解 (...after taking [calcium]_Tr and other treatments, [back pain]_S was significantly alleviated) |
|  | TrWS | Treatment worsen symptom | ...发现[血糖升高]_S，口服[拜糖平]_Tr 及[二甲双胍]_Tr8 天，血糖控制欠佳 (...found that [blood glucose rose]_S, oral [acarbose]_Tr and [metformin]_Tr eight days, poorly controlled blood glucose) |
|  | TrCS | Treatment causes symptom | ...应用长效干扰素[派罗欣]_Tr 后出现[体力下降]_S，[周身不适]_S (...after application of [Pegasys]_Tr, appeared [physical decline]_S and [general malaise]_S) |
|  | TrAS | Treatment is administered for symptom | ...于医院查[肌酐增高]_S，给与患者[改善肾血流]_Tr 等相关治疗 (...checked out [creatinine increased]_S in the hospital, and the patient was given [improvement of renal blood flow]_Tr and other related treatments) |
|  | TrNAS | Treatment is not administered because of symptom | ...发现[转氨酶高]_S，停用[达那唑]_Tr... (...found [high transaminase]_S, stopped taking [danazol]_Tr...) |
| Tests, Diseases | TeRD | Test reveals disease | [头CT]_Te 示：[双侧多发腔梗]_D ([head CT]_Te showed: [bilateral multiple lacunar infarct]_D) |
|  | TeCD | Test conducted to investigate disease | 患者病情尚不除外[脑炎]_D,建议[腰穿]_Te... ([encephalitis]_D was not excepted in the patient's conditions, suggest [lumbar puncture check]_Te...) |
| Tests, Symptoms | TeRS | Test reveals symptom | ...[头CT检查]_Te 显示[颅内多发低密度病灶]_S (...[head CT examination]_Te showed [intracranial multiple low density lesions]_S) |
|  | TeAS | Test is administered because of symptom | ...出现[发热]_S，[鼻出血]_S，当地查[血常规]_Te... (...appeared [fever]_S, [epistaxis]_S, and checked [blood routine]_Te in local...) |
| Diseases, Symptoms | DCS | Disease causes symptom | 3 年前[脑梗死]_D 遗留[说话含糊不清]_S，[走路拖沓]_S... ([cerebral infarction]_D three years ago, now presenting with [muffled speech]_S, [walk procrastination]_S...) |
|  | SID | Symptom indicates disease | ...出现[胸闷]_S[心悸]_S 等不适,诊断[甲亢性心脏病]_D (...had discomforts such as [chest congestion]_S and [palpitation]_S, and was diagnosed with [hyperthyroid heart disease]_D) |

In the examples, entities are in brackets followed by the abbreviation of the entity type. D, diseases; S, symptoms; Te, tests; Tr, treatments.

*3.3. Annotation method*

Referring to the annotation methods in English clinical texts [11, 12], annotation guideline development and corpus construction for each NLP task were executed in three major stages (as shown in Fig. 4):

1. Building the draft guidelines: Annotation guidelines for CTB [35-37] and annotation guidelines in the 2010 i2b2/VA challenge [31-33] were chosen as the basis for developing guidelines for NLP tasks on Chinese clinical texts. With the help of two resident physicians (P1 and P2), four annotators with backgrounds in computational linguistics (CL1 and CL2 for low-level tasks, CL5 and CL6 for higher-level tasks) summarized the characteristics of Chinese clinical texts and drafted annotation guidelines adapted for them. In these guidelines, a large number of annotated examples are listed, and annotation ambiguities are analyzed in detail to make the annotation work easier.
2. Training the annotators and updating the guidelines: An iterative method was proposed to train the annotators and update the guidelines. In each round, a certain number of clinical documents were randomly sampled from the unannotated dataset. To accelerate the annotation progress as well as to ensure annotation quality, different strategies were implemented during the double-annotation period of different tasks: (1) automated tools trained in the general domain [44-46] were applied in the pre-tagging of low-level annotations, and four annotators with backgrounds in computational linguistics were divided into two groups (CL1 and CL3 in annotator group 1, CL2 and CL4 in annotator group 2) to conduct double verification and correction of the automatically added annotations (the annotators in each group accomplish the work collaboratively); annotation disagreements were then adjusted by a physician (P1); (2) since annotations of entities and assertions require professional medical knowledge, we had two physicians (P1 in annotator group 3, P2 in annotator group 4) annotate documents in parallel from the beginning; (3) in the relation annotation task, the documents were double-annotated by two annotator groups (CL5 and CL7 in annotator group 5, CL6 and CL8 in annotator group 6), and a physician (P2) was also assigned to resolve the annotation differences. IAA was then calculated to measure the quality of annotator training, and inconsistent cases were discussed to update the annotation guidelines.
3. Corpus construction: The iterative process in stage 2 continued until IAA was consistently high in the latest three iterations, showing that annotators reached an agreement on annotation guidelines. After the iterative annotator training process, two annotator groups in each task labeled different datasets separately to reduce the consumption of time and money. During this period, three measures were taken to ensure annotation quality: (1) duplicate documents were assigned to two annotator groups for the IAA evaluation of stage 3; (2) annotators recorded uncertain annotations, whose final results were achieved after discussion; (3) sampling inspection was carried out and at least one third of the annotations were checked, and the conflicts with the latest guidelines were then modified after further discussion.

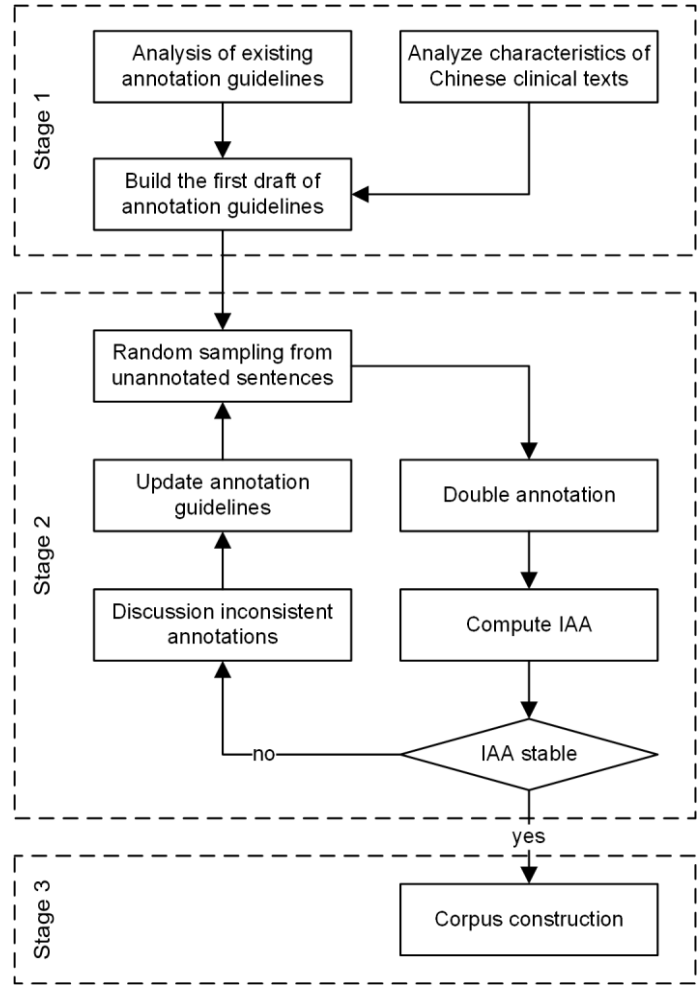

**Fig. 4.** Iterative annotation method for guideline development and corpus construction. IAA, inter-annotator agreement.

*3.4. Inter-annotator agreement*

To evaluate the annotation quality of our corpus, IAA was calculated using the $F_1$ measure. The annotations of one annotator group were seen as the gold standard, and were used to calculate the precision, recall, and $F_1$ measure of the second annotator group, as described in the following equations [47]:

$$Precision = AgreedNumber(AG_1, AG_2) / AnnotationNumber(AG_2), \tag{1}$$

$$Recall = AgreedNumber(AG_1, AG_2) / AnnotationNumber(AG_1), \tag{2}$$

$$F = (1+\beta^2) * Precision * Recall / (\beta^2 * Precision + Recall), \tag{3}$$

where $AgreedNumber(x, y)$ means the number of the consistent annotations between $x$ and $y$, $AnnotationNumber(x)$ means the annotation number of $x$, $AG_i$ means annotator group $i$, and $\beta = 1$ in our work.

For parsing annotations, Evalb [48] was utilized to calculate the IAA of the parsing trees. Since entities and their assertions were annotated simultaneously to accelerate the annotation progress, we merged these two IAA evaluations into one, in which the agreement should satisfy the condition that the extent, type, and assertion of an entity are consistent. Considering the existence of entity groups in entity relations, two types of IAA for relations were computed: the first measured the IAA of relation annotations that preserve entity groups in the relationship; the second separated entity groups into entities and then calculated the IAA of the one-to-one relationships.

**4. Results**

*4.1. Annotation consistency*

As shown in Table 4, the IAA values of these annotation tasks show an increasing trend in the latest three annotator training iterations, indicating that an annotator's mastery of the annotation guidelines improves continually. Furthermore, on account of the fact that the IAA values of relation annotations in the training stage are relatively lower, we added duplicate documents in the corpus construction stage of higher-level tasks. The last column of Table 4 shows that the IAA of these documents remained at a relatively high level, indicating that annotators have the ability to accomplish these annotation tasks with acceptable consistencies.

**Table 4**

Inter-annotator agreement in the latest three annotator training iterations and corpus construction stage ($F_1$ measure)

|  | IAA | | | |
| --- | --- | --- | --- | --- |
|  | Training[-3] | Training[-2] | Training[-1] | Corpus construction |
| Word segmentation | 0.965 | 0.979 | 0.983 | – |
| POS tagging | 0.893 | 0.952 | 0.956 | – |
| Shallow parsing | 0.956 | 0.969 | 0.970 | – |
| Full parsing | 0.805 | 0.840 | 0.865 | – |
| Entity (span, type, assertion) | 0.848 | 0.920 | 0.927 | 0.922 |
| Relation (entity group preserved) | 0.765 | 0.781 | 0.843 | 0.772 |
| Relation (one-to-one) | 0.742 | 0.774 | 0.805 | 0.755 |

"–" means not evaluated. IAA, inter-annotator agreement; POS, part-of-speech.

*4.2. Data analysis of annotations for low-level tasks*

Annotations for low-level tasks cover 72 Chinese discharge summaries and 66 progress notes, including 2553 full parsing trees. There are 47,424 tokens in this corpus, and its average sentence length is 18.58 tokens, which is much shorter than the 27.09 in CTB 5.0. Within clinical texts, the average sentence length of discharge summaries is shorter than that of progress notes (14.13 vs. 22.42) because sentences in some sections of discharge summaries are quite short, especially in the case of only one token in the "treatment effect" section. Fig. 5 gives a detailed comparison between tag distributions in Chinese clinical texts and CTB.

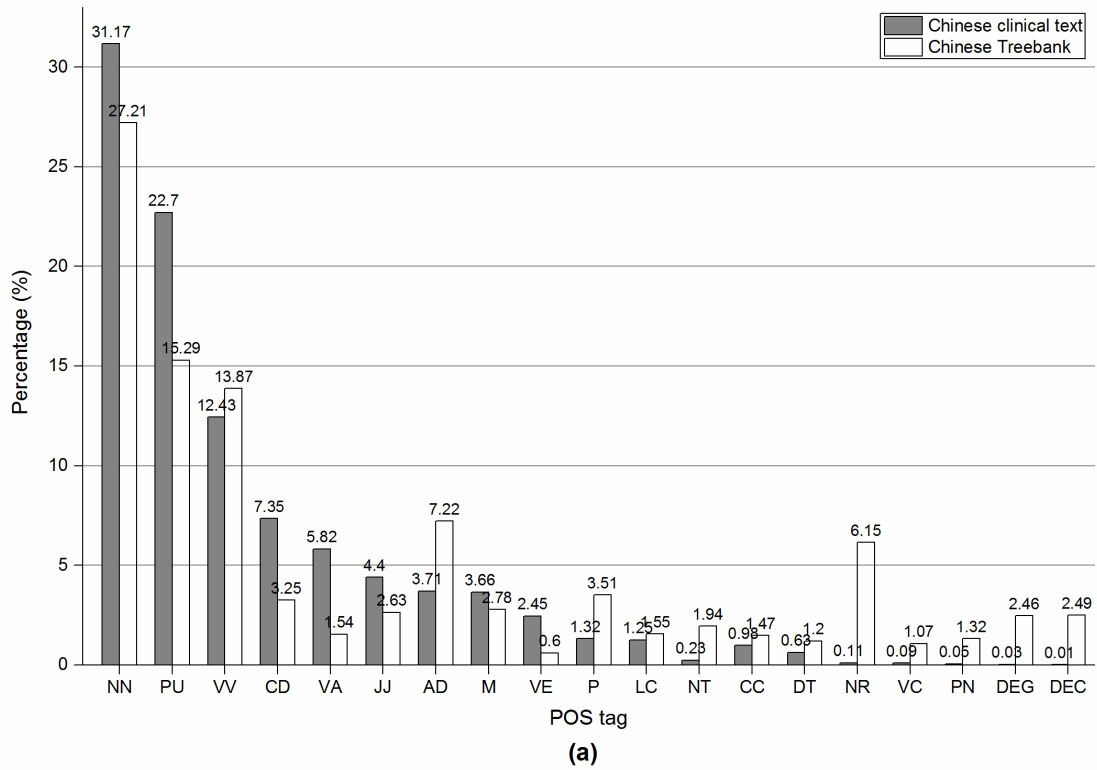
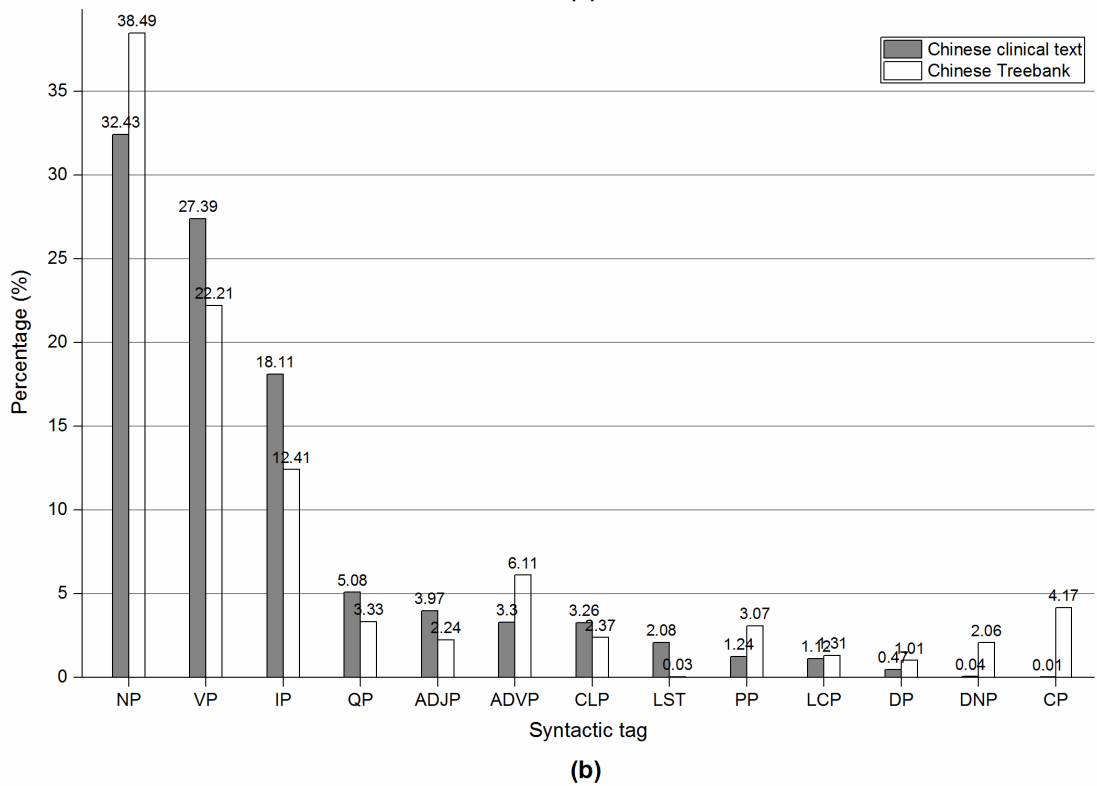

**Fig. 5.** POS and syntactic tag distributions in Chinese clinical texts and CTB 5.0. The tags, whose percentages in clinical texts and CTB are both below 1%, are not listed in this figure. CTB, Chinese Treebank; NN, common nouns; PU, punctuation; VV, other verbs; CD, cardinal numbers; VA, predicative adjective; JJ, noun-modifier other than nouns; AD, adverbs; M, measure word; VE, you3 as the main verb; P, prepositions; LC, localizer; NT, temporal nouns; CC, coordinating conj; DT, determiner; NR, proper nouns; VC, copula shi4; PN, pronouns; DEG, associative de5; DEC, de5

for relative-clause etc.; POS, part-of-speech; NP, noun phrase; VP, verb phrase; IP, simple clause; QP, quantifier phrase; ADJP, adjective phrase; ADVP, adverbial phrase; CLP, classifier phrase; LST, list marker; PP, preposition phrase; CLP, phrase formed by "phrase + LC"; DP, determiner phrase; DNP, phrase formed by "phrase + DEG"; CP, clause headed by complementizer.

Compared with CTB, the POS tag distribution in clinical texts is relatively concentrated, and some tags are rare, such as NR (proper nouns), VC (copula shi4), PN (pronouns), DEG (associative de5), and DEC (de5 for relative-clause etc.). The low percentage of NR in clinical texts is due to the de-identification of patients. Furthermore, the 22.7% of PU (punctuation) in clinical texts is much higher than the 15.29% in CTB because phrase structures, which are separated by punctuations, appear frequently in clinical texts to describe patients' conditions. Moreover, some of the test results in clinical texts are described in the form of a numerical value, resulting in the percentage of CD (cardinal numbers) much higher than that in CTB.

As shown in Fig. 5b, syntactic tag distribution in clinical texts is quite different from that of CTB, and this is closely related to the sublanguage properties of clinical texts. Some syntactic tags are rare in clinical texts, such as DNP (phrase formed by "phrase + DEG") and CP (clause headed by complementizer). Moreover, the low proportion of DNP can be attributed to the same low percentage of DEG in POS tags. Furthermore, some sections in clinical texts, such as case characteristics and treatment plans, are detailed in the form of a list. For this reason, the 2.08% of LST (list marker) in clinical texts is understandably higher than the 0.03% in CTB.

*4.3. Data analysis of annotations for higher-level tasks*

Annotations for higher-level tasks contain 500 discharge summaries and 492 progress notes, including 39,511 entities and 7695 one-to-one relations. Compared with discharge summaries, entities and relations contained in progress notes occur in larger quantities, accounting for three fifths and four fifths of the total numbers, respectively. Fig. 6 shows entity and relation type distributions in these discharge summaries and progress notes.

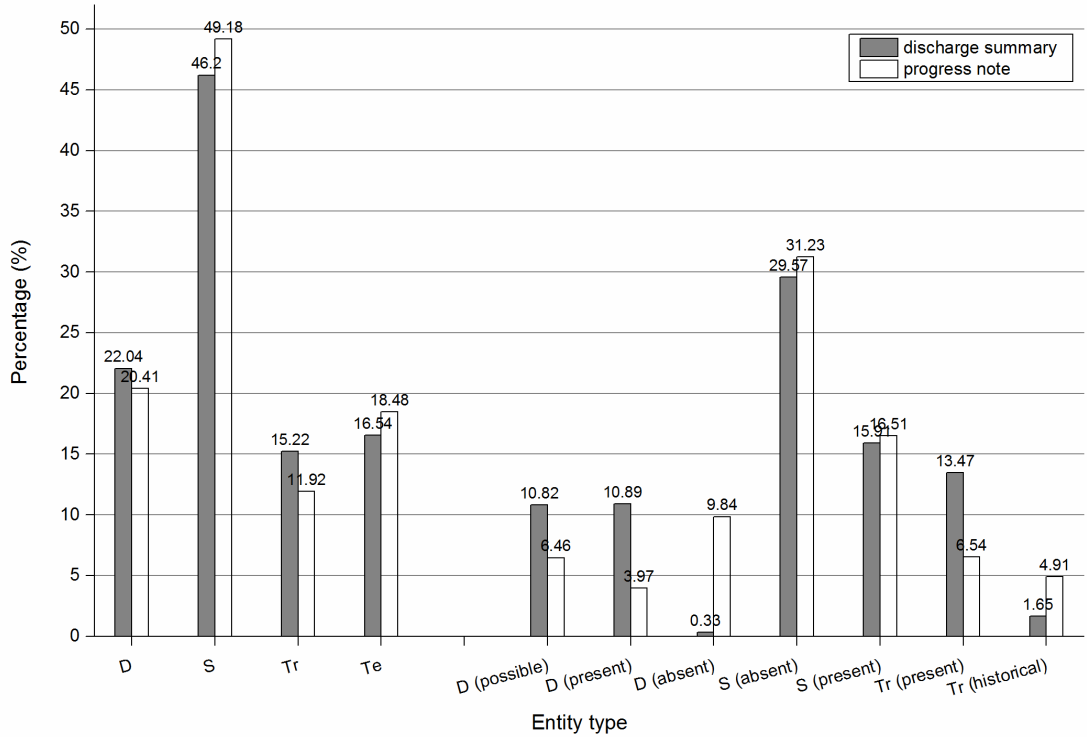

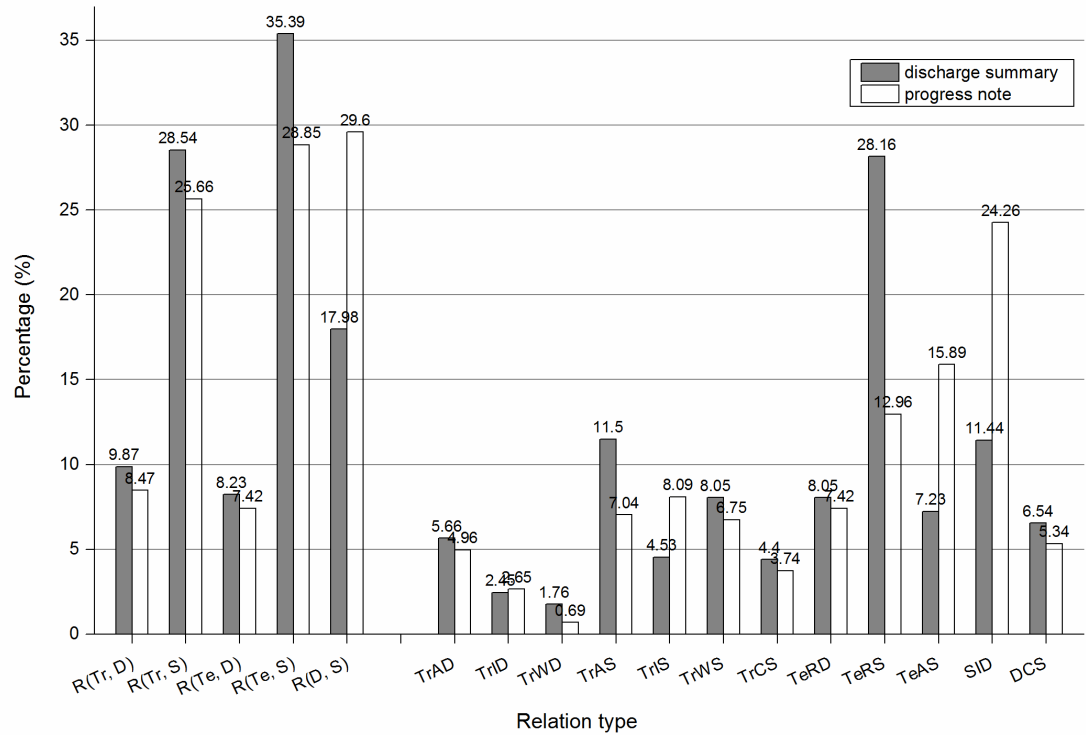

**Fig. 6.** Entity and relation type distributions in Chinese discharge summaries and progress notes. The types, whose percentages in discharge summaries and progress notes are both below 1%, are not listed in this figure. D, diseases; S, symptoms; Tr, treatments, Te, tests; R (entity1, entity2), relation between entity1 and entity2; TrAD, treatment is administered for disease; TrID, treatment improves disease; TrWD, treatment worsen disease; TrAS, treatment is administered for symptom; TrIS, treatment improves symptom; TrWS, treatment worsen symptom; TrCS,

treatment causes symptom; TeRD, test reveals disease; TeRS, test reveals symptom; TeAS, test is administered because of symptom; SID, symptom indicates disease; DCS, disease causes symptom.

Discharge summaries and progress notes have similar distributions of the four entity types, as shown in Fig. 6a. Symptoms account for nearly half of the total entities in discharge summaries and progress notes, respectively, and almost three fifths of these symptoms are *absent*, which can be used by physicians to distinguish patients' conditions. In addition to these approximate distributions, the proportions of some assertion types in discharge summaries and progress notes show some differences. In discharge summaries, admission diagnosis results in more *possible* diseases, while clinical definite diagnosis leads to more *present* diseases; however, case characteristics describe the patient's medical history, leading to many more *absent* diseases and *historical* treatments in progress notes.

In Fig. 6b, relation type distributions in discharge summaries and progress notes are quite different for some relation types, especially disease-symptom relations, and this is closely related to the content emphasis of different clinical text types. In progress notes, present illness history is presented in the section of case characteristics, including patients' conditions, tests, and relevant diagnoses, so the proportion of TeAS (test is administered because of symptom) and SID (symptom indicates disease) are much higher than those in discharge summaries.

*4.4. System development*

To verify the usefulness of our annotated corpus, we developed a chinese clinical text processing and information extraction system (CCTPIES) that consisted of a word segmenter, POS tagger, shallow parser, full parser, named entity recognizer, and relation extractor [49], and the performance of these modules was evaluated by 10-fold cross validation on the annotated corpus; results are shown in Table 5.

**Table 5**

Performance of system modules trained on our annotated clinical texts

| Module | Precision | Recall | F1 |
| --- | --- | --- | --- |
| Word segmenter | 0.981 | 0.979 | 0.980 |
| POS tagger | 0.966 | 0.964 | 0.965 |
| Shallow parser | 0.946 | 0.949 | 0.948 |
| Full parser | 0.845 | 0.841 | 0.843 |
| Named entity recognizer | 0.923 | 0.902 | 0.912 |
| Relation extractor | 0.784 | 0.691 | 0.735 |

We used a sequence-labeling method to train statistical models for word segmentation, POS tagging, shallow parsing, and named entity recognition. CRF++ [50], an open-source implementation of the conditional random fields algorithm, was used to train these models. As shown in Table 5, the evaluation results of these modules trained by CRF++ are quite excellent in that all of them achieved the level of practical application. To build a full parsing model, we trained the Stanford parser and the Berkeley parser [51] on our annotated corpus; results showed that both parsers were satisfactory, but that the Berkeley parser was slightly better. However, there were some null outputs in the Berkeley parser, so we used the corresponding outputs in the Stanford parser to replace them. This improvement further enhanced the evaluation of the full parser, and we chose this combined parser as our full parser. Moreover, similarly to most relation extraction research on English clinical texts, we used the support vector machines (SVM) algorithm to train models on our annotated Chinese clinical texts, and LIBSVM [52] was selected as the training tool.

**5. Discussion**

*5.1. Contributions of this work*

In this study, we constructed a comprehensive syntactic and semantic corpus of Chinese clinical texts, covering annotations for word segmentation, POS tagging, shallow parsing, full parsing, NER, assertion classification, and relation extraction.

Because extensive medical knowledge exists in clinical texts, we referred to annotation guidelines from the general domain and the clinical domain, and developed annotation guidelines for Chinese clinical texts with the help of physicians. As described in the guideline development section, many improvements were proposed to adapt to the characteristics of Chinese clinical texts.

Before building the corpus, annotators kept training by following annotation guidelines until their annotation consistency remained at a relatively high level. During the annotation period, existing open-source tools were used for pre-labeling, and significantly reduced the burden on annotators.

As is widely known, double annotation improves corpus quality; however, as the corpus scale grows, annotation costs in terms of time and money can be a challenge. Therefore, we balanced these factors and proposed an annotation method: double annotation was adopted in the annotator training stage; then, annotators were allowed to annotate separately in the corpus construction stage, using certain annotation quality assurance measures. The annotation consistency shows that our annotated corpus is of good quality.

Based on this corpus, some syntactic and semantic features of Chinese clinical texts were analyzed (more analysis results are provided in Appendix A). Moreover, a Chinese clinical text processing and information extraction system was developed, and its modules can be seen as baselines for research in the clinical domain. To our knowledge, some of these modules described here are introduced into Chinese texts in the clinical domain for the first time, including the POS tagger, shallow parser, full parser.

*5.2. Limitations and future work*

Although our annotated corpus makes a contribution to research on Chinese texts in the clinical domain, there are some limitations in our study. Because of limited annotation resources, the syntactic corpus only covers two departments within the Second Affiliated Hospital of Harbin Medical University. There are differences in medical terminologies from different hospital departments, which may weaken the adaptability of some NLP techniques across different departments.

As future work, some explorations will be conducted. First, transfer learning approaches will be introduced to solve the adaptation problem among different hospital departments. Second, some additional types of clinical text should be annotated to improve the practicability of NLP techniques developed based on this corpus. Third, active learning methods will be explored to reduce the annotation burden on annotators by filtering redundant samples from unlabeled data while selecting undertrained samples for the annotators. Finally, algorithms used to improve the performance of NLP systems for clinical texts will be developed.

**6. Conclusions**

In this paper, we described the construction of a corpus of Chinese clinical texts using an iterative annotation method. By following the annotation guidelines developed in this study, good levels of annotation consistency were achieved. Moreover, a CCTPIES was developed to verify the usefulness of the corpus, which achieved excellent performance. To the best of our knowledge, this corpus is the first comprehensive annotated corpus of Chinese texts in the clinical domain, laying a solid foundation for future research. The related annotation resources are available at http://github.com/WILAB-HIT/Resources.

**Author contributions**

This work was a collaboration of all the authors. BH, JY, ZJ, and CQ developed the annotation guidelines and took part in corpus construction. BH, BD, YG, and QY performed corpus analysis. BH, ZJ, and JC developed system modules and evaluated their performance. All authors contributed to drafting, revision, and final approval of this manuscript.

**Funding**

This work was supported by the Ricoh Software Research Center (Beijing).

**Conflict of interest**

The authors have no conflicts of interest to declare.

**Acknowledgements**

We would like to thank the medical records room of the Second Affiliated Hospital of Harbin Medical University for providing the Chinese clinical texts used in this study. We are grateful to our physicians, YZ and YL, and the following annotators with backgrounds in computational linguistics: FZ, XD, TQ, and YP.

**Appendix A. Distributions of annotations in Chinese clinical texts**

Part A (Part-of-speech tags in the syntactic corpus)

| Annotation type | Description | Counts | % in our annotated corpus |
|---|---|---|---|
| NN | common nouns | 14,782 | 31.17 |
| PU | punctuation | 10,763 | 22.70 |
| VV | other verbs | 5896 | 12.43 |
| CD | cardinal numbers | 3484 | 7.35 |
| VA | predicative adjective | 2762 | 5.82 |
| JJ | noun-modifier other than nouns | 2086 | 4.40 |
| AD | adverbs | 1759 | 3.71 |
| M | measure word (including classifiers) | 1736 | 3.66 |
| VE | you3 as the main verb | 1160 | 2.45 |
| P | prepositions (excluding ba3 and bei4) | 628 | 1.32 |
| LC | localizer | 595 | 1.25 |
| NT | temporal nouns | 584 | 1.23 |
| CC | coordinating conj | 463 | 0.98 |
| DT | determiner | 251 | 0.53 |
| OD | ordinal numbers | 232 | 0.49 |
| ETC | tags for deng3 and deng3deng3 in coordination phrases | 74 | 0.16 |
| NR | proper nouns | 53 | 0.11 |
| VC | copula shi4 | 44 | 0.09 |
| PN | pronouns | 26 | 0.05 |
| DEG | associative de5 | 16 | 0.03 |
| MSP | some particles | 8 | 0.02 |
| CS | subordinating conj | 7 | 0.01 |
| DEC | de5 for relative-clause etc. | 6 | 0.01 |
| SB | bei4 in short bei-construction | 5 | 0.01 |
| BA | ba3 in ba-const | 1 | <0.01 |
| FW | foreign words | 1 | <0.01 |
| LB | bei4 in long bei-construction | 1 | <0.01 |
| AS | aspect marker | 1 | <0.01 |
| SP | sentence-final particle | 0 | 0 |
| DER | de5 in V-de const. and V-de-R | 0 | 0 |
| DEV | de5 as the head of DVP | 0 | 0 |
| IJ | interjection | 0 | 0 |
| ON | onomatopoeia | 0 | 0 |

Part B (Syntactic tags in the syntactic corpus)

| Annotation type | Description | Counts | % in our annotated corpus |
|---|---|---|---|
| NP | noun phrase | 17,254 | 32.43 |
| VP | verb phrase | 14,573 | 27.39 |
| IP | simple clause | 9634 | 18.11 |
| QP | quantifier phrase | 2701 | 5.08 |
| ADJP | adjective phrase | 2114 | 3.97 |
| ADVP | adverbial phrase | 1754 | 3.30 |
| CLP | classifier phrase | 1736 | 3.26 |
| LST | list marker | 1104 | 2.07 |
| PP | preposition phrase | 662 | 1.24 |
| LCP | phrase formed by "phrase + LC" | 598 | 1.12 |
| FRAG | fragment | 341 | 0.64 |
| DP | determiner phrase | 251 | 0.47 |
| VCD | coordinated verb compound | 164 | 0.31 |
| VSB | verb compounds formed by a modifier + a head | 121 | 0.23 |
| PRN | parenthetical | 106 | 0.20 |
| VRD | verb resultative compound | 37 | 0.07 |

| | | | | |
|---|---|---|---|---|
| UCP | unidentical coordination phrase | | 28 | 0.05 |
| DNP | phrase formed by "phrase + DEG" | | 23 | 0.04 |
| CP | clause headed by C (complementizer) | | 6 | 0.01 |
| VPT | potential form V-de-R or V-bu-R | | 1 | <0.01 |
| VNV | verb compounds formed by A-not-A or A-one-A | | 1 | <0.01 |
| VCP | verb compounds formed by VV + VC | | 1 | <0.01 |
| DVP | phrase formed by "phrase + DEV" | | 0 | 0 |

Part C (Entities and assertions in the semantic corpus)

| Annotation type | Counts | % in the corresponding entity type | % in all the annotated entities |
|---|---|---|---|
| Diseases: Possible | 3255 | 39.09 | 8.24 |
| Diseases: Present | 2686 | 32.25 | 6.80 |
| Diseases: Absent | 2352 | 28.24 | 5.95 |
| Diseases: Not associated with the patient | 35 | 0.42 | 0.09 |
| Diseases: Conditional | 0 | 0.00 | 0.00 |
| Diseases: Occasional | 0 | 0.00 | 0.00 |
| *Diseases: Total* | *8328* | *100.00* | *21.08* |
| | | | |
| Symptoms: Absent | 12,070 | 63.69 | 30.55 |
| Symptoms: Present | 6425 | 33.90 | 16.26 |
| Symptoms: Conditional | 257 | 1.36 | 0.65 |
| Symptoms: Occasional | 153 | 0.81 | 0.39 |
| Symptoms: Possible | 41 | 0.22 | 0.10 |
| Symptoms: Not associated with the patient | 5 | 0.03 | 0.01 |
| *Symptoms: Total* | *18,951* | *100.00* | *47.96* |
| | | | |
| Treatments: Present | 3703 | 70.63 | 9.37 |
| Treatments: Historical | 1413 | 26.95 | 3.58 |
| Treatments: Absent | 127 | 2.42 | 0.32 |
| *Treatments: Total* | *5243* | *100.00* | *13.27* |
| | | | |
| *Tests: Total* | *6989* | *100.00* | *17.69* |

Part D (Relations in the semantic corpus)

| Annotation type | Description | Counts | % in the corresponding entity pair | % in all the annotated relations |
|---|---|---|---|---|
| TrAD | Treatment is administered for disease | 393 | 58.66 | 5.11 |
| TrID | Treatment improves disease | 201 | 30.00 | 2.61 |
| TrWD | Treatment worsen disease | 70 | 10.45 | 0.91 |
| TrCD | Treatment causes disease | 6 | 0.90 | 0.08 |
| *R(Tr, D)* | | *670* | *100.00* | *8.71* |
| TrAS | Treatment is administered for symptom | 613 | 30.35 | 7.97 |
| TrIS | Treatment improves symptom | 566 | 28.02 | 7.36 |
| TrWS | Treatment worsen symptom | 540 | 26.73 | 7.02 |
| TrCS | Treatment causes symptom | 298 | 14.75 | 3.87 |
| TrNAS | Treatment is not administered because of symptom | 3 | 0.15 | 0.04 |
| *R(Tr, S)* | | *2020* | *100.00* | *26.26* |
| | | | | |
| TeRD | Test reveals disease | 581 | 99.49 | 7.55 |
| TeCD | Test conducted to investigate disease | 3 | 0.51 | 0.04 |
| *R(Te, D)* | | *584* | *100.00* | *7.59* |
| | | | | |
| TeRS | Test reveals symptom | 1239 | 53.31 | 16.11 |
| TeAS | Test is administered because of symptom | 1085 | 46.69 | 14.11 |
| *R(Te, S)* | | *2324* | *100.00* | *30.22* |
| | | | | |
| SID | Symptom indicates disease | 1663 | 79.46 | 21.62 |
| DCS | Disease causes symptom | 430 | 20.54 | 5.59 |
| *R(D, S)* | | *2093* | *100.00* | *27.21* |

R(entity1, entity2), relation between entity1 and entity2; D, diseases; S, symptoms; Te, tests; Tr, treatments.

Part E (Part-of-speech tags in the syntactic corpus: discharge summary vs progress note)

| Annotation type | Description | % in discharge summaries | % in progress notes |
|---|---|---|---|
| NN | common nouns | 32.90 | 30.23 |
| PU | punctuation | 21.29 | 23.46 |
| VV | other verbs | 12.85 | 12.20 |
| CD | cardinal numbers | 6.86 | 7.61 |
| VA | predicative adjective | 6.62 | 5.39 |
| JJ | noun-modifier other than nouns | 4.41 | 4.39 |
| AD | adverbs | 3.40 | 3.88 |
| M | measure word (including classifiers) | 3.71 | 3.63 |
| VE | you3 as the main verb | 2.09 | 2.64 |
| P | prepositions (excluding ba3 and bei4) | 0.86 | 1.58 |
| LC | localizer | 0.93 | 1.43 |
| NT | temporal nouns | 1.84 | 0.90 |
| CC | coordinating conj | 0.74 | 1.11 |
| DT | determiner | 0.54 | 0.52 |
| OD | ordinal numbers | 0.81 | 0.31 |
| ETC | tags for deng3 and deng3deng3 in coordination phrases | 0.09 | 0.19 |
| NR | proper nouns | 0 | 0.17 |
| VC | copula shi4 | 0.02 | 0.13 |
| PN | pronouns | 0.02 | 0.07 |
| DEG | associative de5 | 0 | 0.05 |
| MSP | some particles | <0.01 | 0.02 |
| CS | subordinating conj | 0.02 | <0.01 |
| DEC | de5 for relative-clause etc. | 0 | 0.02 |
| SB | bei4 in short bei-construction | 0 | 0.02 |
| BA | ba3 in ba-const | 0 | <0.01 |
| FW | foreign words | 0 | <0.01 |
| LB | bei4 in long bei-construction | 0 | <0.01 |
| AS | aspect marker | 0 | <0.01 |
| SP | sentence-final particle | 0 | 0 |
| DER | de5 in V-de const. and V-de-R | 0 | 0 |
| DEV | de5 as the head of DVP | 0 | 0 |
| IJ | interjection | 0 | 0 |
| ON | onomatopoeia | 0 | 0 |

Part F (Syntactic tags in the syntactic corpus: discharge summary vs progress note)

| Annotation type | Description | % in discharge summaries | % in progress notes |
|---|---|---|---|
| NP | noun phrase | 33.27 | 31.95 |
| VP | verb phrase | 27.38 | 27.39 |
| IP | simple clause | 18.08 | 18.12 |
| QP | quantifier phrase | 5.17 | 5.02 |
| ADJP | adjective phrase | 3.90 | 4.01 |
| ADVP | adverbial phrase | 2.95 | 3.49 |
| CLP | classifier phrase | 3.23 | 3.28 |
| LST | list marker | 1.60 | 2.34 |
| PP | preposition phrase | 0.83 | 1.48 |
| LCP | phrase formed by "phrase + LC" | 0.78 | 1.32 |
| FRAG | fragment | 1.45 | 0.18 |
| DP | determiner phrase | 0.47 | 0.47 |
| VCD | coordinated verb compound | 0.50 | 0.20 |
| VSB | verb compounds formed by a modifier + a head | 0.22 | 0.23 |
| PRN | parenthetical | 0.07 | 0.27 |
| VRD | verb resultative compound | 0.05 | 0.08 |
| UCP | unidentical coordination phrase | 0.03 | 0.06 |
| DNP | phrase formed by "phrase + DEG" | 0.01 | 0.06 |
| CP | clause headed by C (complementizer) | 0.01 | 0.01 |
| VPT | potential form V-de-R or V-bu-R | 0 | <0.01 |
| VNV | verb compounds formed by A-not-A or A-one-A | 0 | <0.01 |
| VCP | verb compounds formed by VV + VC | 0 | <0.01 |
| DVP | phrase formed by "phrase + DEV" | 0 | 0 |